\documentclass{article}
\usepackage{spconf,amsmath,graphicx}

\usepackage{amssymb}
\usepackage{booktabs}

\usepackage[inline]{enumitem}

\usepackage{multirow}
\usepackage{makecell}
\usepackage{xcolor}
\usepackage{pifont} 
\newcommand{\cmark}{\ding{51}}
\newcommand{\xmark}{\ding{55}}

\usepackage{caption}
\usepackage{subcaption}

\usepackage{url}

\DeclareMathOperator*{\argmin}{arg\,min}

\title{Diversifying Human Pose in Synthetic Data for Aerial-view Human Detection}
\name{Yi-Ting~Shen$^{\star,1}$~~~
Hyungtae~Lee$^{\star,2}$~~~
Heesung~Kwon$^{3}$~~~ Shuvra~S.~Bhattacharyya$^{1}$\thanks{$^{\star}$Contributed equally.}}
\address{$^{1}$University of Maryland, College Park~~~
$^{2}$BlueHalo~~~
$^{3}$DEVCOM Army Research Laboratory}
\begin{document}
%
\maketitle
\begin{abstract}
  Synthetic data generation has emerged as a promising solution to the data scarcity issue in aerial-view human detection. However, creating datasets that accurately reflect varying real-world human appearances—particularly diverse poses—remains challenging and labor-intensive. To address this, we propose SynPoseDiv, a novel framework that diversifies human poses within existing synthetic datasets. SynPoseDiv tackles two key challenges: generating realistic, diverse 3D human poses using a diffusion-based pose generator, and producing images of virtual characters in novel poses through a source-to-target image translator. The framework incrementally transitions characters into new poses using optimized pose sequences identified via Dijkstra’s algorithm. Experiments demonstrate that SynPoseDiv significantly improves detection accuracy across multiple aerial-view human detection benchmarks, especially in low-shot scenarios, and remains effective regardless of the training approach or dataset size.
\end{abstract}

\section{Introduction}
\label{sec:intro}

Despite the recent surge in the use of unmanned aerial vehicles (UAVs), aerial scene perception, particularly  object detection in aerial images, has not progressed as much as general object detection~\cite{zou2023object}. This gap may be attributed to the lack of real-world datasets that sufficiently capture diverse object appearances, especially humans, from aerial perspectives. 

Leveraging synthetic data---generated by manipulating factors related to human appearances---for training detectors has recently emerged as a promising solution~\cite{YShenCVPR2023}. Efforts have also been made to synthesize images of human postures using multiple virtual characters with varied outfits and physical characteristics~\cite{MBlackCVPR2023}, or by capturing them from different angles and distances~\cite{YShenAccess2023}. However, constructing a synthetic dataset with diverse human appearances that match real-world characteristics typically requires significant efforts from experts to model characters and design motions. Human poses, in particular, are a key factor contributing to distinctive human appearances; they are highly diverse and only partially defined, making them exceptionally difficult to control. This challenge becomes even more pronounced when everything must be built from scratch, a common scenario due to the fact that existing synthetic datasets often do not share the complete assets (e.g., characters, motions, or backgrounds) used to generate them.

\begin{figure}[t]
\centering
\includegraphics[trim=0mm 0mm 0mm 0mm,clip,width=\linewidth]{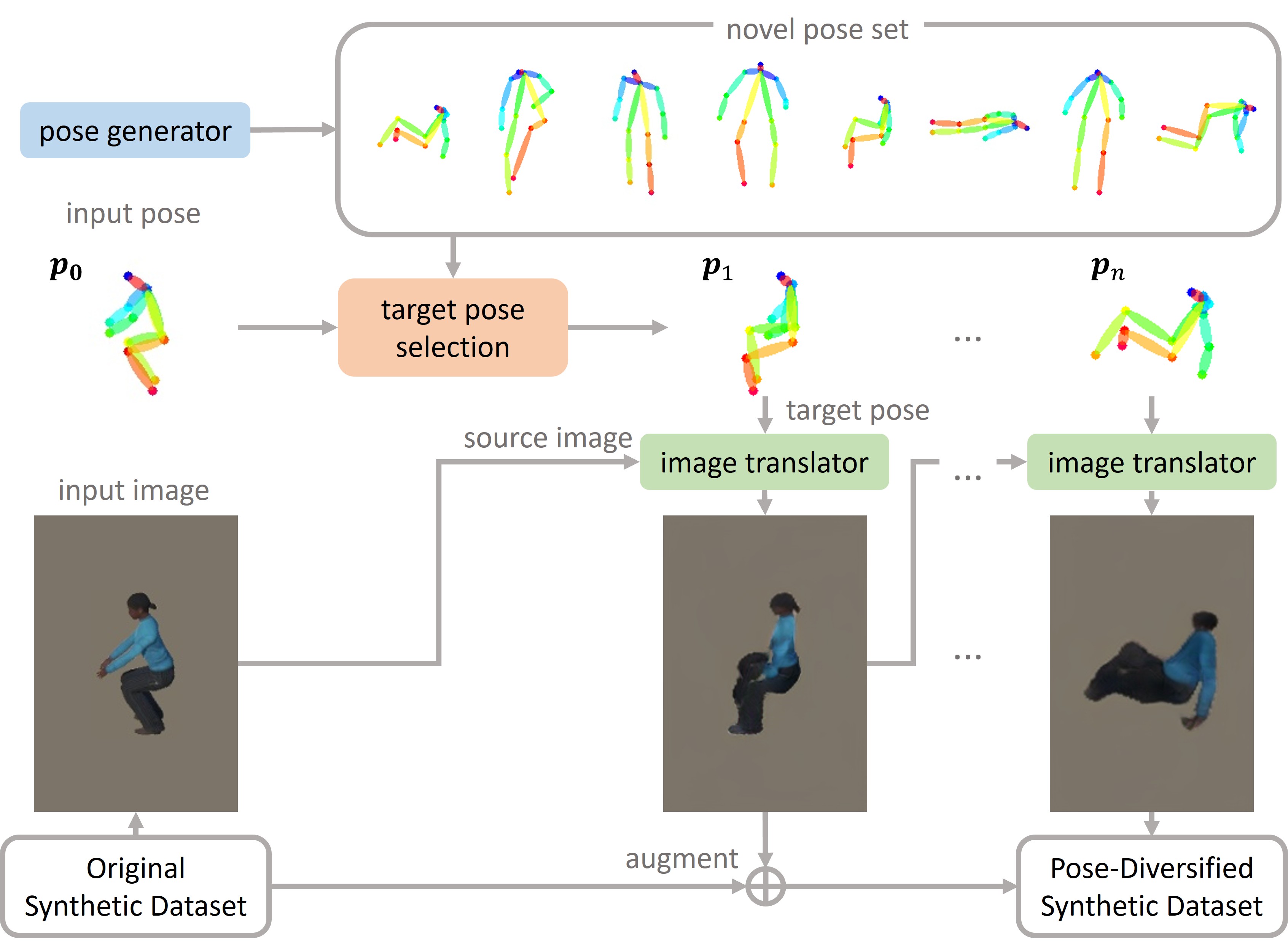}
\caption{{\bf The SynPoseDiv framework} generates a set of novel poses using a pose generator. For each synthetic image in the original dataset, a sequence of target poses is selected from the novel pose set through target pose selection. The image translator then generates images corresponding to each target pose in the sequence. A pose-diversified synthetic dataset is constructed by adding the generated images to the original synthetic dataset after post-processing.}
\label{fig:flowchart}
\end{figure}

In this paper, we introduce an alternative approach called SynPoseDiv---a novel framework that diversifies human poses in an existing synthetic dataset (Fig.~\ref{fig:flowchart}). Specifically, SynPoseDiv transforms images in the existing dataset so that the characters assume poses not found in the original set, while maintaining the original style. To achieve this, SynPoseDiv addresses two key challenges: i) \emph{how to identify a variety of novel yet realistic poses}, and ii) \emph{how to generate human images in these novel poses}.

To tackle the first challenge, SynPoseDiv incorporates a pose generator based on the diffusion model, an emerging generative model known for its remarkable performance~\cite{GTevetICLR2023}. The pose generator produces a wide range of realistic human poses in the form of 3D keypoint locations. Using this generator, SynPoseDiv constructs a large set of diverse novel poses (`novel pose set') while ensuring diversity by filtering out highly similar poses.

To address the second challenge, SynPoseDiv employs a source-to-target image translator to create human images where virtual characters assume various poses from the novel pose set. Since target images corresponding to these novel poses are unavailable for training---which could limit the image translator's effectiveness---SynPoseDiv applies the translator to sequences of target poses (`target pose sequences') that demonstrate \emph{incremental} body movements, gradually transforming an input pose into the desired final target pose. These target pose sequences are selected from the novel pose set through a target pose selection process, which identifies the optimal path from a starting pose to a final target pose within a graph built from the novel pose set. This process is solved using Dijkstra's algorithm~\cite{EDijkstraNM1959}. For each target pose in the sequence, a corresponding target image is generated by the image translator, using the previously generated target image as the source.

We demonstrate that leveraging SynPoseDiv to diversify a dataset remarkably enhances detection performance, especially in low-shot scenarios where additional data is crucial. Moreover, regardless of the training approach (either `pre-training and fine-tuning' or `Progressive Transformation Learning (PTL)~\cite{YShenCVPR2023}'), the pose-diversified synthetic dataset consistently yields significantly better detection accuracy than the original synthetic dataset across three aerial-view human detection benchmarks: VisDrone~\cite{PZhuTPAMI2022}, Okutama-Action~\cite{MBarekatainCVPRW2017}, and ICG~\cite{ICGlink}. The positive impact of using the pose-diversified synthetic dataset persists regardless of the dataset size, as supported by a series of scaling behavior analyses from various perspectives. Additionally, SynPoseDiv is designed to be general and not limited by the network architecture of the pose generator or image translator, allowing it to benefit from any future advancements in these areas.
\section{Related Works}
\label{sec:rel_works}

SynPoseDiv offers a novel framework unifying human image synthesis and pose generation to enhance detector performance. We briefly outline related areas:\smallskip

\noindent{\bf Synthesizing human appearances.} Prior work has manipulated clothing~\cite{MBlackCVPR2023} and viewing angles~\cite{YShenAccess2023} to synthesize data using parameterized sampling. However, the vast diversity of human poses makes selecting representative samples challenging. To address this, some methods~\cite{CIonescuTPAMI2014} use MoCap in controlled settings to capture real poses, while others synthesize poses from 2D images using 3D data like depth~\cite{VGabeurICCV2019}. Similar to our goal, prior work has modified existing images to depict varied poses. For instance, HUSC~\cite{MZanfirAAAI2020} used an image translator to create unseen poses. In this paper, we introduce an image translator that tackles the same challenge, generating novel poses without ground truth images.\smallskip

\noindent{\bf Pose generation.} Prior work has generated novel poses by completing missing keypoints using visible joints (e.g., transformers~\cite{ZGengCVPR2023}) or sampling from a pose prior (e.g., multi-task representations~\cite{HCiCVPR2023}). Additionally, adversarial learning~\cite{ADavydovCVPR2022} has been used to guide pose generators to produce plausible poses. Buliding on the second approach, we developed a diffusion model-based pose generator.\smallskip

\noindent{\bf Generating images of novel poses.} Many efforts have focused on transforming human images to match target poses using generators like conditional GANs~\cite{NLiICCV2023} and diffusion models~\cite{XHanICCV2023}, consistently improving generation quality. To address the challenge of training without paired source and target pose images, DPE~\cite{YPangCVPR2023} used a bidirectional cyclic training strategy that minimized differences between images of the same character after aligning pose and expression. However, this required optimal face reconstruction, which is difficult in our case. To overcome this, we train an image translator with limited transformation capability, applying it only when source and target poses are similar. We construct the target pose sequence with sufficiently similar adjacent poses and generate target images in sequence.
\section{The SynPoseDiv Framework}
\label{sec:method}

\begin{figure}[t]
\centering
\includegraphics[trim=0mm 600mm 1000mm 0mm,clip,width=\linewidth]{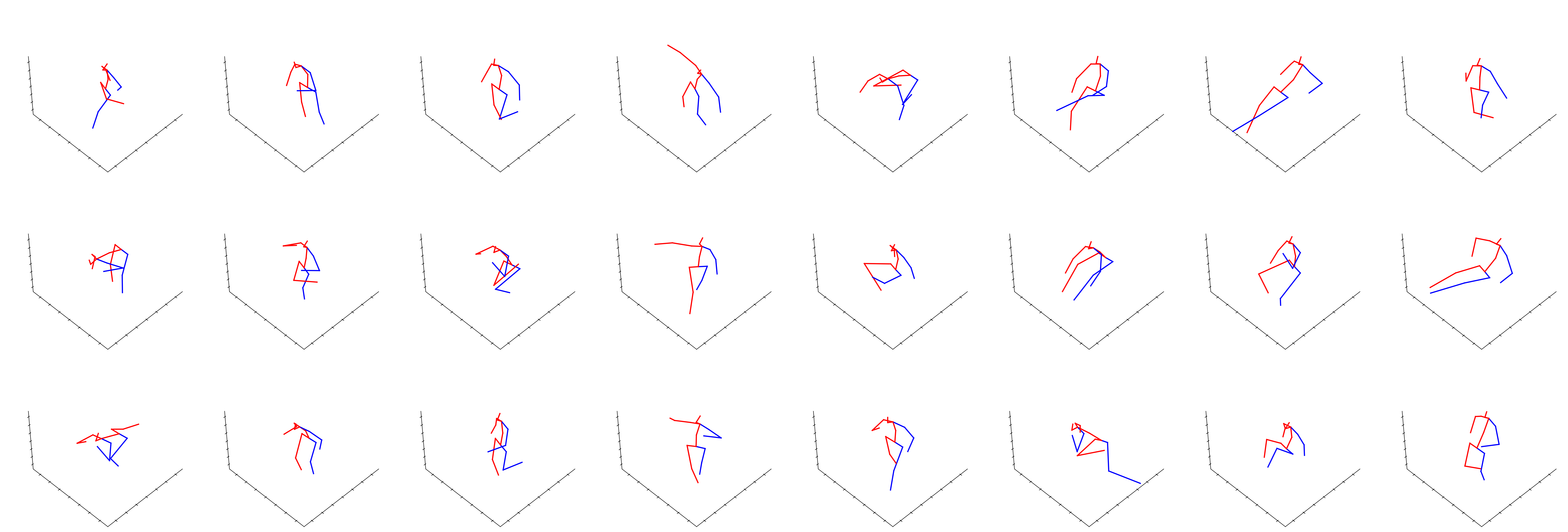}
\caption{{\bf Sample poses in $\mathcal{P}_n$.} Poses are represented by 3D locations of 17 keypoints.}
\label{fig:novel_poses}
\end{figure}

\noindent{\bf Pose generator.} The pose generator is designed to create novel yet realistic poses absent from existing datasets. Each generated pose is represented as 3D coordinates of $K$ keypoints, denoted ${\bf p}\in \mathbb{R}^{K\times 3}$ (where $K=17$ in our case). To achieve this, the pose generator employs a diffusion model, an emerging generative model known for its impressive performance in various domains~\cite{GTevetICLR2023}. Using this generator, we construct a novel pose set $\mathcal{P}_n$. To ensure diversity within the set, we exclude pose pairs that exhibit high similarity. Specifically, a newly generated pose is added to the set only if it is dissimilar to existing poses, defined by a pose distance (eq.~\ref{eq:pdist}) less than a specified threshold $T_\text{sim}$ (in our case, $T_\text{sim}=0.24$). This process continues until $N_\text{pos}$ poses are included in the set. Fig.~\ref{fig:novel_poses} presents examples of poses in $\mathcal{P}_n$.\smallskip

\begin{figure}
\centering
\includegraphics[trim=0mm 0mm 0mm 0mm,clip,width=\linewidth]{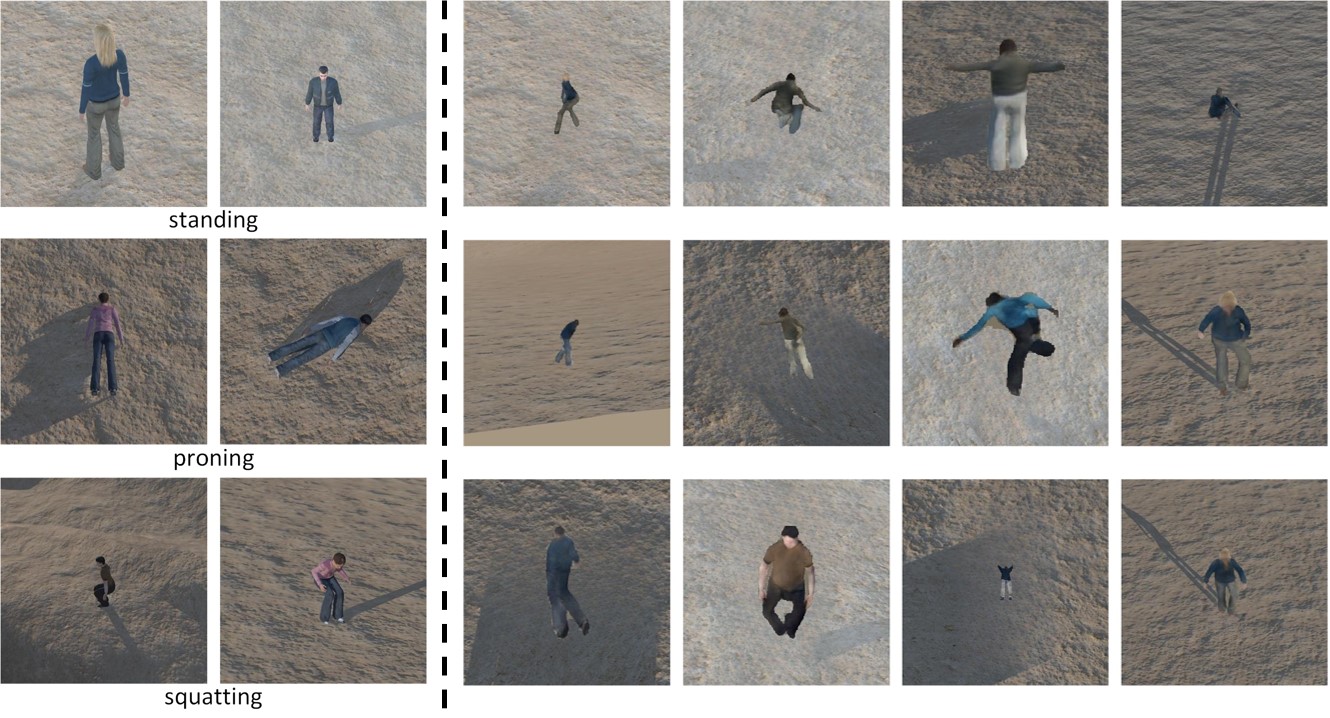}
\caption{{\bf Sample generated images} before applying post-processing are shown on the right. Images of the three poses from the original synthetic dataset~\cite{YShenAccess2023} are shown on the left.}
\label{fig:novel_images}
\end{figure}

\noindent{\bf Target pose selection.} Due to the lack of target images corresponding to novel poses for training, the image translator may struggle to generate images with poses significantly different from the input. To address this limitation, we introduce \textit{progressive pose search} that evolves sequential target poses from the input pose. This method leverages Dijkstra's algorithm~\cite{EDijkstraNM1959} to search for the optimal path from the start node (input pose) to the destination node (target pose) within a graph representation of the novel pose set $\mathcal{P}_n$. The input pose is estimated as 2D human keypoints using ViTPose~\cite{YXuNeurIPS2022} on the input image, while the target pose is randomly selected from $\mathcal{P}_n$. Intermediate target poses ${\bf p}_1,~{\bf p}_2,~\cdots,~{\bf p}_{n-1}$ between the input pose (${\bf p}_0$) and the final target pose (${\bf p}_n$) are optimally determined from $\mathcal{P}_n$ by minimizing the objective:
\begin{equation}
d({\bf p}_0,{\bf p}_n)=\sum_{i=1}^n\left(d_\text{pose}\left({\bf p}_{i-1}, {\bf p}_i\right)\right)^k~~~\text{s.t.},|\mathcal{P}_t|\leq N_\text{max},\label{eq:obj}
\end{equation}
where $d_\text{pose}$ denotes the pose distance (eq.~\ref{eq:pdist}) between the 2D keypoint locations of two poses, and $\mathcal{P}_t=\{{\bf p}_1,~{\bf p}_2,~\cdots,~{\bf p}_n\}$ represents the target pose set. To reduce cumulative generation errors, we limit the maximum number of target poses to $N_\text{max}$. In our setup, $k=2$ ensures that shorter pose distances have a greater impact on the objective.
Dijkstra's algorithm is solved using \textit{dynamic programming}, with eq.~\ref{eq:obj} expressed as recurrence relations:
\begin{eqnarray}
d\left({\bf p}_{i-1}, {\bf p}_n\right)=\min_{{\bf p}\in\mathcal{P}_n}{\left(\left(d_\text{pose}\left({\bf p}_{i-1},{\bf p}\right)\right)^k+d\left({\bf p}, {\bf p}_n\right)\right)},\label{eq::sub_optim_d}\\
{\bf p}_i=\argmin_{{\bf p}\in\mathcal{P}_n}{\left(\left(d_\text{pose}\left({\bf p}_{i-1},{\bf p}\right)\right)^k+d\left({\bf p}, {\bf p}_n\right)\right)}.\label{eq:sub_optim_p}
\end{eqnarray}
The dynamic programming approach decomposes the optimization into sub-problems from ${\bf p}_1$ to ${\bf p}_{n-1}$, which are solved recursively in reverse order (i.e., optimally finding ${\bf p}_{n-1}$ to ${\bf p}_1$). This recursive formulation mirrors the shortest-path computation in Dijkstra’s algorithm, where each intermediate pose is chosen to minimize the cumulative pose distance to the final target pose. Once a target pose sequence is constructed via progressive pose search, the image translator is applied to each target pose sequentially, using the previously generated image as the source for the next image generation.\smallskip

\noindent{\bf Image translator.} Training the image translator requires source poses, target poses, and their corresponding images. Since target images corresponding to novel poses are unavailable for training, we fine-tune a pre-trained image translator~\cite{YRenCVPR2022} using pose pairs from the original synthetic dataset~\cite{YShenAccess2023}, rather than relying on the novel poses in $\mathcal{P}_n$. After fine-tuning, the image translator generates human images in significantly different poses from the input by iterative application. The translator is applied to human regions segmented from the input synthetic image to prevent non-human regions from adversely affecting generation. Segmentation masks, typically available with synthetic data, are used for this purpose. The segmented human region in the input image is then filled using background inpainting~\cite{RSuvorovWACV2022}. The generated human images are resized at random ratios and positioned at random locations within background images, following the human-size distribution of the original synthetic dataset. Fig.~\ref{fig:novel_images} illustrates examples of generated images.\smallskip

\begin{figure}[t]
\centering
\setlength{\tabcolsep}{5pt}
\begin{tabular}{cccc}
\includegraphics[trim=5mm 5mm 5mm 5mm,clip,width=.29\linewidth]{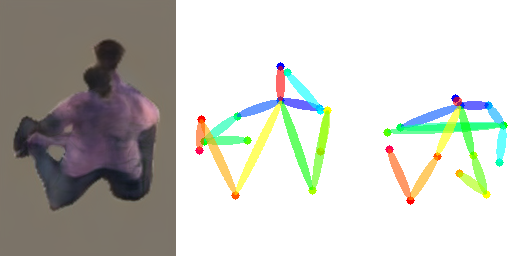} &
\includegraphics[trim=5mm 5mm 5mm 5mm,clip,width=.29\linewidth]{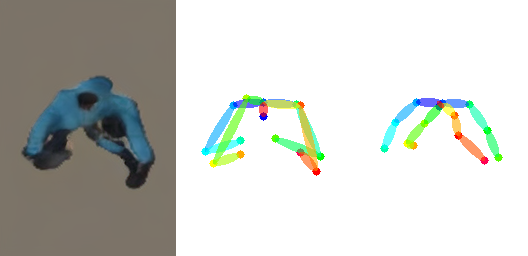} &
\includegraphics[trim=5mm 5mm 5mm 5mm,clip,width=.29\linewidth]{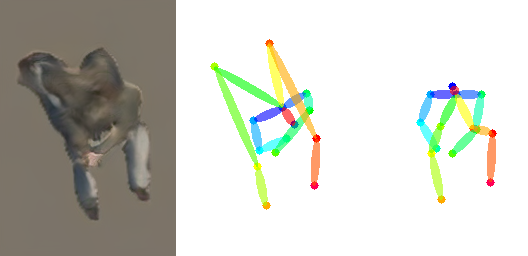}\\
$d_\text{pose}=0.32$ & $d_\text{pose}=0.16$ & $d_\text{pose}=0.22$ \\
\end{tabular}
\caption{{\bf Filtered noisy images.} Each example shows a generated image (left), a target pose (middle), and an estimated pose (right). The pose distance $d_\text{pose}$ is displayed beneath.}
\label{fig:noisy_images}
\end{figure}

\begin{figure}[t]
\centering
\resizebox{\linewidth}{!}{%
\setlength{\tabcolsep}{2.0pt}
\renewcommand{\arraystretch}{1.2}
\begin{tabular}{cccccc}
\includegraphics[trim=0mm 0mm 0mm 0mm,clip,width=.2\linewidth]{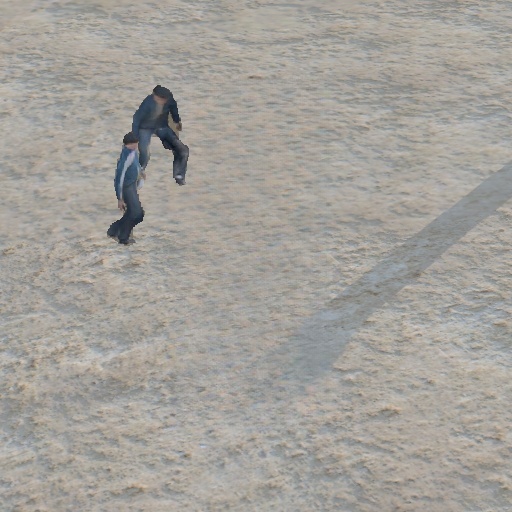} 
&
\includegraphics[trim=0mm 0mm 0mm 0mm,clip,width=.2\linewidth]{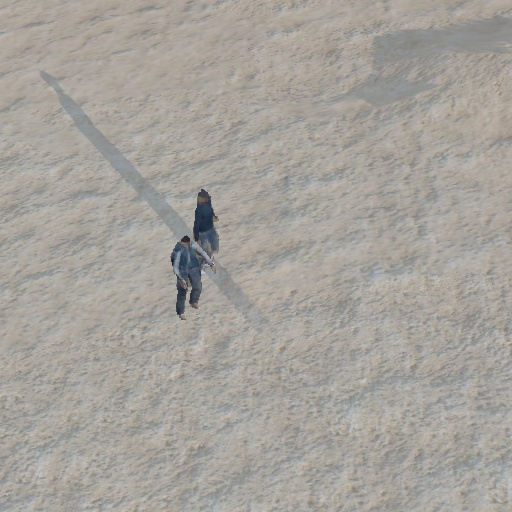}
&
\includegraphics[trim=0mm 0mm 0mm 0mm,clip,width=.2\linewidth]{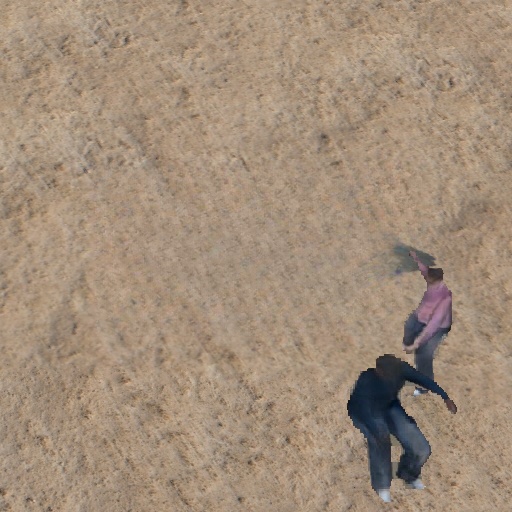}
&
\includegraphics[trim=0mm 0mm 0mm 0mm,clip,width=.2\linewidth]{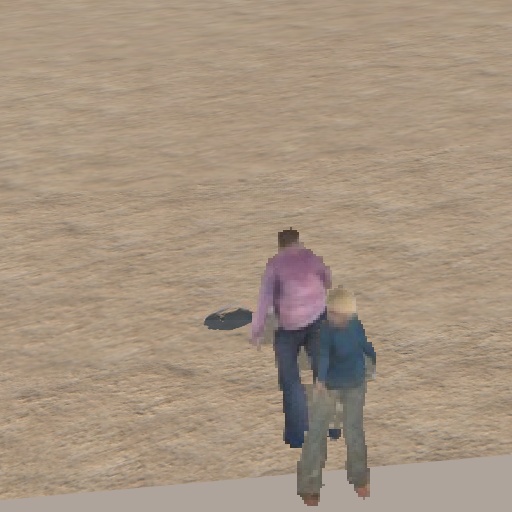}
&
\includegraphics[trim=0mm 0mm 0mm 0mm,clip,width=.2\linewidth]{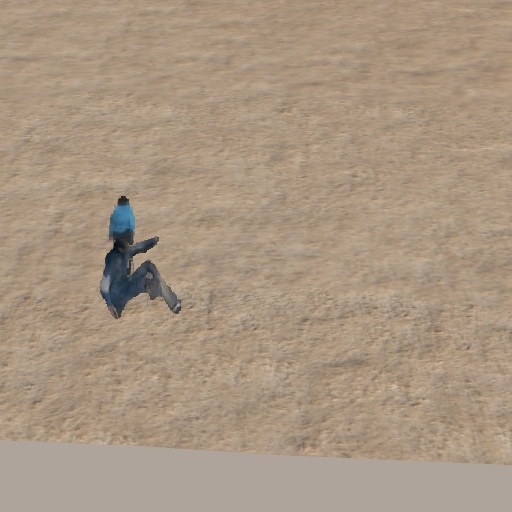}
&
\includegraphics[trim=0mm 0mm 0mm 0mm,clip,width=.2\linewidth]{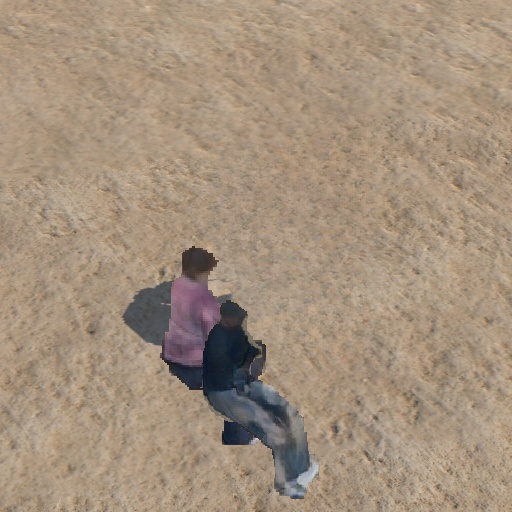}
\end{tabular}
}
\caption{{\bf Sample generated images} after occlusion insertion.}
\label{fig:novel_img_samples}
\end{figure}

\noindent{\bf Post-processing: noisy image filtering and occlusion insertion.} We begin by filtering out noisy images that could negatively impact performance. We conduct a sanity check based on pose estimation to identify and remove low-quality images from all generated images. If the estimated poses differ significantly from the target poses used during generation, these images are considered noisy. Specifically, images are filtered out if the pose distances (eq.~\ref{eq:pdist}) between the target and estimated poses exceeds a threshold $T_\text{filt}$. In our setup, $T_\text{filt}=0.1$. Examples of noisy generated images are shown in Fig.~\ref{fig:noisy_images}.

Next, we address occlusions~\cite{ansarian2021realistic}, which significantly contribute to variations in human appearances in real-world aerial-view images. We focus on a common scenario where one human occludes another, often observed in aerial-view images. To simulate this, two generated humans with the same camera position are placed within a single image. Their locations are randomly determined to ensure overlap, with the occluding human positioned lower than the occluded one to mimic natural occlusions in aerial views. Examples of generated images after this step are shown in Fig.~\ref{fig:novel_img_samples}.

\section{Implementation Details}
\label{ssec:implementation_detail}

\noindent{\bf Measuring pose similarity.} The pose distance $d_\text{pose}$, used in several aspects of our approach, is calculated by comparing the keypoint locations (2D or 3D) of two poses ${\bf p}_i$ and ${\bf p}_j$ as follows: 
\begin{equation}
d_\text{pose}({\bf p}_i, {\bf p}_j)=\sqrt{2\left(1-{\bar {\bf p}_i}^\top {\bar {\bf p}_j^\text{tr}}\right)},\label{eq:pdist}
\end{equation}
where ${\bar {\bf p}_i}$ and ${\bar {\bf p}_j^\text{tr}}$ are $L_2$ normalized pose vectors. ${\bf p}_j^\text{tr}$ represents a transformed ${\bf p}_j$ adjusted such that both input poses share the same facing direction and are captured from the same camera position. To determine the facing direction of each pose, we use the direction of the \textit{neck-to-nose} vector. This alignment is generally effective in our case, where head and neck movements are not dominant. The camera position for each pose is either provided by the original synthetic dataset as metadata or is easily obtainable. When comparing a 2D pose and a 3D pose (as used in `target pose selection'), the pose similarity is calculated after transforming the 3D pose and projecting it into the 2D space.\smallskip

\noindent{\bf Pose generator.} Inspired by \cite{GTevetICLR2023}, the pose generator utilizes a diffusion model with two reverse, iterative processes: i) \textit{diffusion} and ii) \textit{denoising}. In the diffusion process at step $t$, the model diffuses the denoised pose ${\bf p}^{(0)}$ into a noisy pose ${\bf p}^{(t)}$ by repeatedly adding Gaussian noise $z_t\sim\mathcal{N}\left({\bf 0}, {\bf I}\right)$ to the input pose for $t$ times. During the denoising process at step $t$, a model $g_\text{pos}$ estimates a clean pose $\hat{{\bf p}}^{(0)}$ from the noisy pose ${\bf p}^{(t)}$. The initial input ${\bf p}^{(t)}$ is drawn from $\mathcal{N}({\bf 0}, {\bf I})$. The model $g_\text{pos}$ employs an 8-layer transformer encoder. Each keypoint is treated as an input token and embedded into a 512-dimensional space via a feed-forward network. The noise time step $t$ is similarly projected using a separate network. Each embedding is combined with a positional embedding and fed to the encoder. Excluding the first output token (representing the noise time step), the remaining tokens are processed through another feed-forward network to predict clean keypoints. The training objective is formulated as:
\begin{equation}
\mathcal{L}({\bf p}^{(t)})=\mathbb{E}_{{\bf p}^{(0)}\sim q({\bf p}^{(0)}),t\sim [1,T]}||{\bf p}^{(0)} - g_\text{pos}({\bf p}^{(t)}, t) ||_2^2.
\end{equation}

\noindent{\bf Image translator.} We employ the pre-trained NTED~\cite{YRenCVPR2022} as the image translator, but SynPoseDiv is easily adaptable to other image translators. The image translator is fine-tuned on Archangel-Synthetic~\cite{YShenAccess2023}, the original synthetic dataset used in work, which contains human image pairs with limited types of similar poses. Image pairs with similar camera positions---information provided as metadata by Archangel-Synthetic---are defined as having similar poses.
\section{Experiments}
\label{sec:experiments}

\noindent{\bf Tasks and datasets.} Following \cite{YShenCVPR2023}, we evaluate aerial-view human detection in two data-scarce scenarios: \textit{low-shot} and \textit{cross-domain} setups, where synthetic data is crucial. Experiments are conducted using the official implementation of \cite{YShenCVPR2023}, training a RetinaNet~\cite{TLinCVPR2017} variant. We use VisDrone~\cite{PZhuTPAMI2022}, Okutama-Action~\cite{MBarekatainCVPRW2017}, and ICG~\cite{ICGlink} as real aerial-view datasets, while Archangel-Synthetic~\cite{YShenAccess2023} serves as the original synthetic dataset. For low-shot setups, we train on 20 or 50 randomly selected VisDrone images (Vis-\{20, 50\}) and evaluate on the VisDrone test set. For cross-domain setups, we test on Okutama-Action and ICG. Results are averaged over three runs to reduce randomness. We use COCO-style AP (i.e., AP$_\text{50:5:95}$) and AP$_\text{50}$ as the evaluation metrics.\smallskip

\noindent{\bf Details of the pose-diversified synthetic dataset.} The pose generator is trained on Human3.6M~\cite{CIonescuTPAMI2014}, utilizing the 17 defined in this dataset, of which 14 are used in the image generator. The novel pose set $\mathcal{P}_n$ contains 1000 poses ($N_\text{pos}=1000$). For each synthetic image, 5 final target poses are selected from $\mathcal{P}_n$, with the maximum number of poses in a target pose sequence ($N_\text{max}$) set to 3. Based on these specifications, the total number of generated novel images is 17,714. The entire generation process took approximately 3 hours.\smallskip

\noindent{\bf Leveraging synthetic data in model training.} We adopt two approaches: pre-training and fine-tuning (`PT-FT'), and progressive transformation learning (PTL)~\cite{YShenCVPR2023}. 'PT-FT' is the most widely used transfer learning method, where a model is initially pre-trained on a synthetic dataset and subsequently fine-tuned on a real dataset. `PTL', a recently proposed method, progressively expands the training set by iteratively selecting subsets of synthetic data. Each selected subset undergoes transformations to enhance realism before being added to the training set. Unless specified otherwise, PTL accuracies reported in this paper are based on results after the 5$^\text{th}$ PTL iteration. In~\cite{YShenCVPR2023}, these two methods were found to be the most effective among various approaches for leveraging synthetic data in training.\smallskip


\begin{table*}[t]
\caption{(a) Detection Accuracy Comparison (b) Effect of Post-Processing (Same-Domain, Vis-20)}
\vspace{-0.2cm}
\begin{subtable}[h]{0.64\textwidth}
\caption{}
\label{tab:main_results}
\centering
\resizebox{\textwidth}{!}{%
\setlength{\tabcolsep}{3.0pt}
\renewcommand{\arraystretch}{1.1}
\begin{tabular}{ll|ccc|ccc}
& & \multicolumn{3}{c|}{Vis-20} & \multicolumn{3}{c}{Vis-50}\\
method & \multicolumn{1}{c|}{set}  & \multicolumn{1}{c}{VisDrone} & \multicolumn{1}{c}{Okutama} & \multicolumn{1}{c|}{ICG} & \multicolumn{1}{c}{VisDrone} & \multicolumn{1}{c}{Okutama} & \multicolumn{1}{c}{ICG} \\\Xhline{1.2pt}
& \textcolor{gray}{real} & \textcolor{gray}{~~0.58/~~~2.27} & \textcolor{gray}{~~3.64/~~~14.54} & \textcolor{gray}{~~0.62/~~~1.89} & \textcolor{gray}{~~0.76/~~~3.30} & \textcolor{gray}{~~~~7.82/~~~28.66} & \textcolor{gray}{~~1.30/~~~5.65} \\\hline
\multirow{2}{*}{PT-FT} & ~~+ orig & ~~0.76/~~~2.48 & ~~4.24/~~~17.17 & ~~6.53/~~23.67 & ~~1.29/~~~3.76 & ~~~~5.32/~~~20.96 & ~~7.10/~~27.95 \\
& ~~+ {\bf p-div} & ~~{\bf 1.21}/~~~{\bf 4.02} & ~~{\bf 9.14}/~~~{\bf 34.70} & ~~{\bf 8.20}/~~{\bf 28.80} & ~~{\bf 1.84}/~~~{\bf 5.37} & ~~~{\bf 10.39}/~~~{\bf 36.83} & ~~{\bf 8.63}/~~{\bf 30.09} \\
& & \textcolor{teal}{(+0.45/+1.54)} & \textcolor{teal}{(+4.90/+17.53)} & \textcolor{teal}{(+1.67/+5.13)} & \textcolor{teal}{(+0.55/+1.61)} & \textcolor{teal}{~(+5.07/+15.87)} & \textcolor{teal}{(+1.53/+2.14)} \\\hline
\multirow{2}{*}{PTL} & ~~+ orig & ~~2.07/~~~6.72 & ~~7.90/~~~31.53 & ~~{\bf 8.81}/~~{\bf 33.71} & ~~2.92/~~~9.26 & ~~~11.49/~~~42.51 & ~~{\bf 8.98}/~~{\bf 33.21} \\
& ~~+ {\bf p-div} & ~~{\bf 2.26}/~~~{\bf 7.39} & ~~{\bf 8.95}/~~~{\bf 36.97} & ~~6.45/~~26.13 & ~~{\bf 2.99}/~~~{\bf 9.42} & ~~~{\bf 12.89}/~~~{\bf 47.24} & ~~6.29/~~25.50\\
& & \textcolor{teal}{(+0.19/+0.67)} & \textcolor{teal}{(+1.05/~~+5.44)} & \textcolor{red}{~(-2.36/-7.58)} & \textcolor{teal}{(+0.07/+0.16)} & \textcolor{teal}{~(+1.40/~+4.73)} & \textcolor{red}{~(-2.69/-7.71)} \\
\end{tabular}
}
\end{subtable}
\hfill
\begin{subtable}[h]{0.34\textwidth}
\caption{}
\label{tab:ablation}
\centering
\resizebox{\linewidth}{!}{%
\setlength{\tabcolsep}{5.5pt}
\renewcommand{\arraystretch}{1.2}
\begin{tabular}{cc|cccc}
\multicolumn{2}{c|}{post-proc.} &&& \multicolumn{2}{c}{acc.$\uparrow$} \\
filt. & occ. & \# img & FID$\downarrow$ & PT-FT & PTL \\\Xhline{1.2pt}
\xmark & \xmark & 34,403 & 124.53 & ~0.89/~2.83 & ~1.90/~6.14 \\
\textcolor{red}{\cmark} & \xmark & 17,714 & {\bf 116.17} & ~0.94/~3.03 & ~2.12/~6.46 \\
\xmark & \textcolor{red}{\cmark} & 34,403 & 121.71 & ~1.11/~3.98 & ~1.95/~6.61 \\
\textcolor{red}{\cmark} & \textcolor{red}{\cmark} & 17,714 & 116.41 & ~{\bf 1.21}/~{\bf 4.02} & ~{\bf 2.26}/~{\bf 7.39} \\
\end{tabular}
}
\end{subtable}
\end{table*}

\noindent{\bf Main results.} In Table~\ref{tab:main_results}, we compare detection accuracy using three training setups: the original synthetic dataset (`real + orig'), the pose-diversified synthetic dataset (`real + p-div'), and only real data (`real'). In nearly all cases---except one (Vis-50 tested on the Okutama-Action dataset)---training with synthetic data improves accuracy over using only real images. Even in the exception, the pose-diversified dataset still enhances accuracy, highlighting the benefit of synthetic data in improving detection performance.

The pose-diversified dataset generally outperforms the original synthetic dataset in boosting detection accuracy, regardless of the training approach. This improvement is more noticeable in cross-domain setups (Okutama-Action and ICG), where the VisDrone training set lacks representative features of the test set. However, in rare cases (PTL on the ICG test set), accuracy declines despite using pose-diversified data, which is discussed later. 

In addition, PTL typically achieves better accuracy than PT-FT, proving more effective at leveraging synthetic data. However, this advantage disappears on the ICG test set. PTL selects synthetic images that are closer in domain to the training set, but due to the significant perspective differences between the VisDrone (oblique) and ICG (nadir) datasets, these images fail to represent ICG data accurately, reducing performance. Conversely, PT-FT, which uses the entire pose-diversified dataset, provides better representation and performance in both same-domain and cross-domain tasks.\smallskip


\noindent{\bf Ablation study on post-processing.} Table~\ref{tab:ablation} examines the impact of post-processing. Additionally, we report FID (Fr{\' e}nchet Inception Distance)~\cite{MHeuselNeurIPS2017}, which quantifies the fidelity and diversity of generated images, using the VisDrone training set as the reference real dataset. The results show that filtering noisy images lowers FID, improving the quality of pose-diversified synthetic data. This improvement in image quality also boosts detection accuracy. Conversely, accounting for occlusions does not reduce FID but still contributes to higher detection accuracy.\smallskip


\begin{figure}[t]
\centering
\setlength{\tabcolsep}{0.5pt}
\begin{tabular}{ccc}
\includegraphics[width=.32\linewidth]{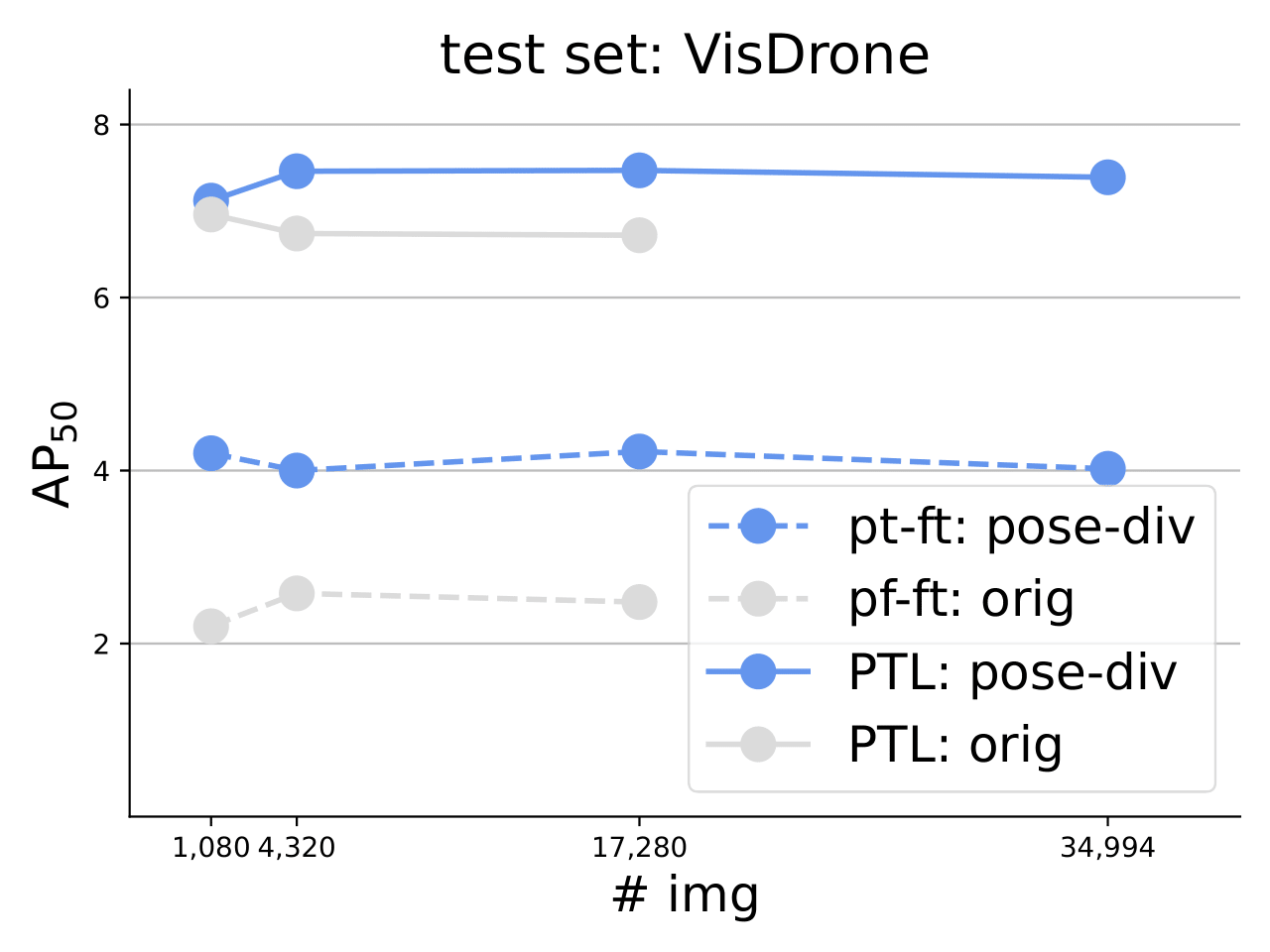} &
\includegraphics[width=.32\linewidth]{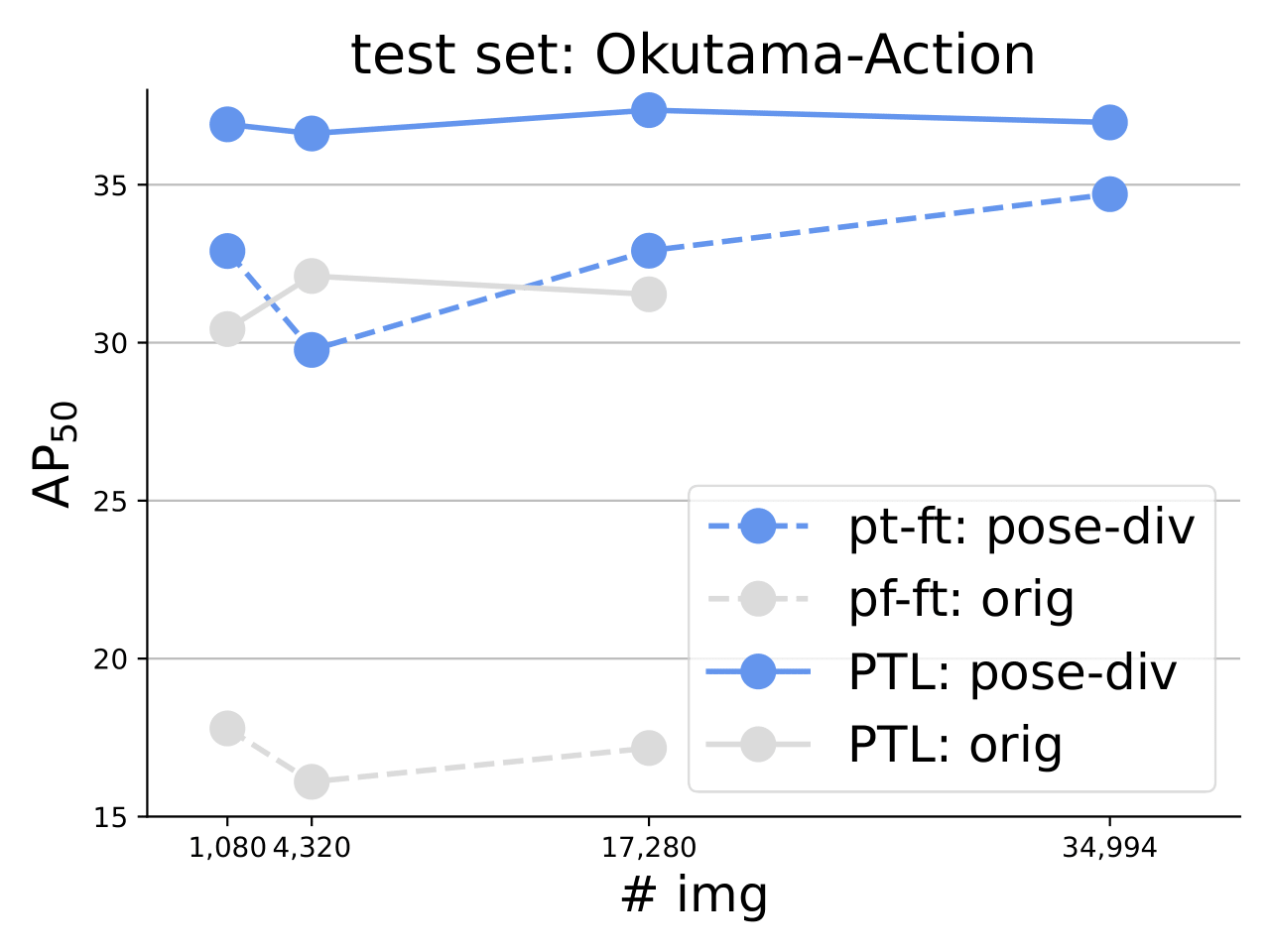} &
\includegraphics[width=.32\linewidth]{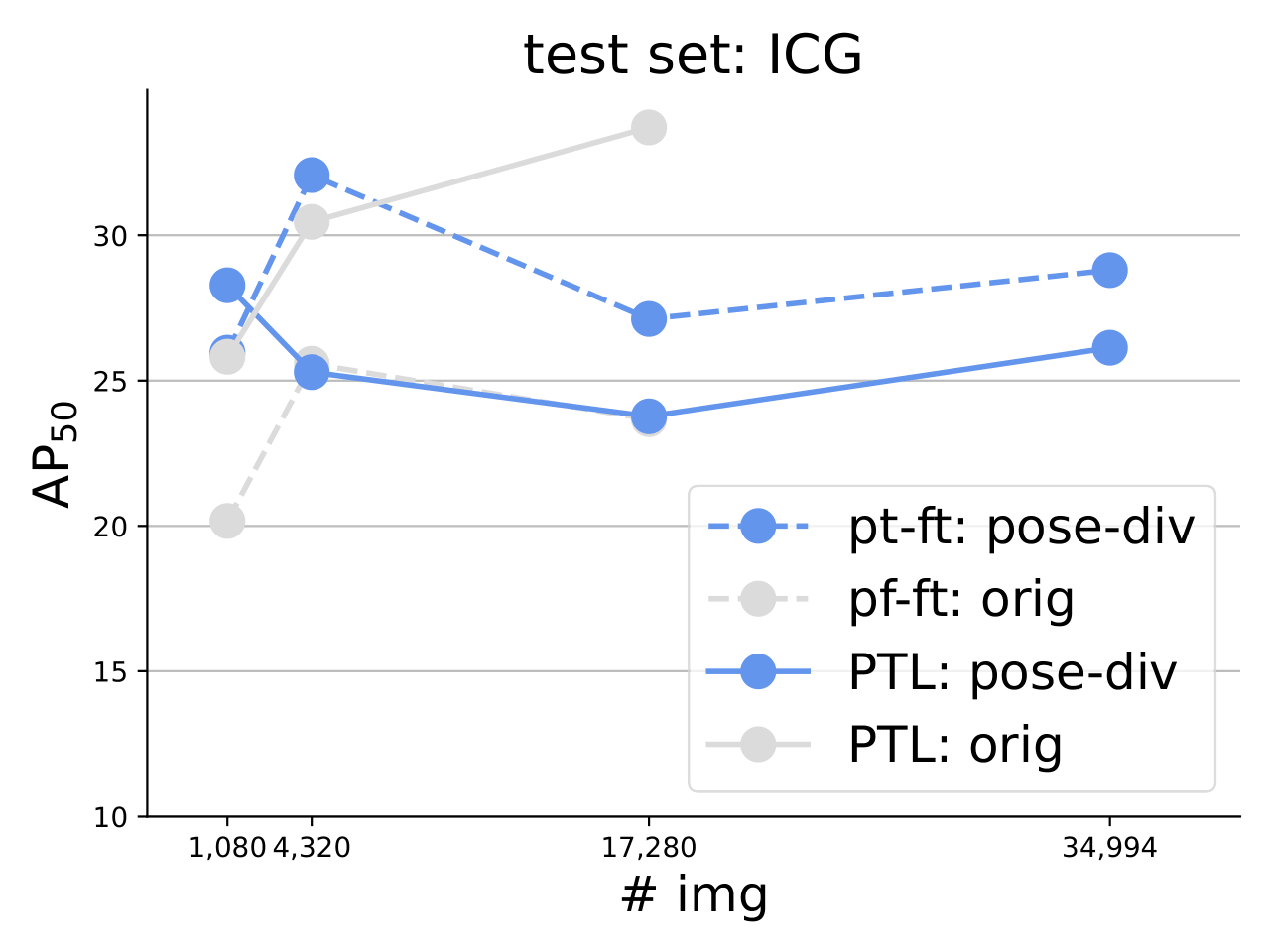} \\
\end{tabular}
\vspace{-0.2cm}
\caption{{\bf Scaling behavior of the synthetic dataset} under the Vis-20 setup. The scaling behavior of each dataset is compared by using randomly sampled subsets of 1,080 (1/16th), 4,320 (1/4th), and 17,280 (full) images.}
\label{fig:scalability_set_size}
\end{figure}

\noindent{\bf Scaling behavior with data size.} One might assume that the improved performance of the pose-diversified synthetic dataset is due to its larger size. However, Fig.~\ref{fig:scalability_set_size} shows it consistently achieves better accuracy on VisDrone and Okutama-Action, even when matched in size with the original synthetic dataset. On ICG, PTL accuracy remains lower with the pose-diversified dataset, consistent with earlier results. This behavior is discussed in later sections.\smallskip

\begin{figure}[t]
\centering
\setlength{\tabcolsep}{0.5pt}
\begin{tabular}{ccc}
\includegraphics[width=.32\linewidth]{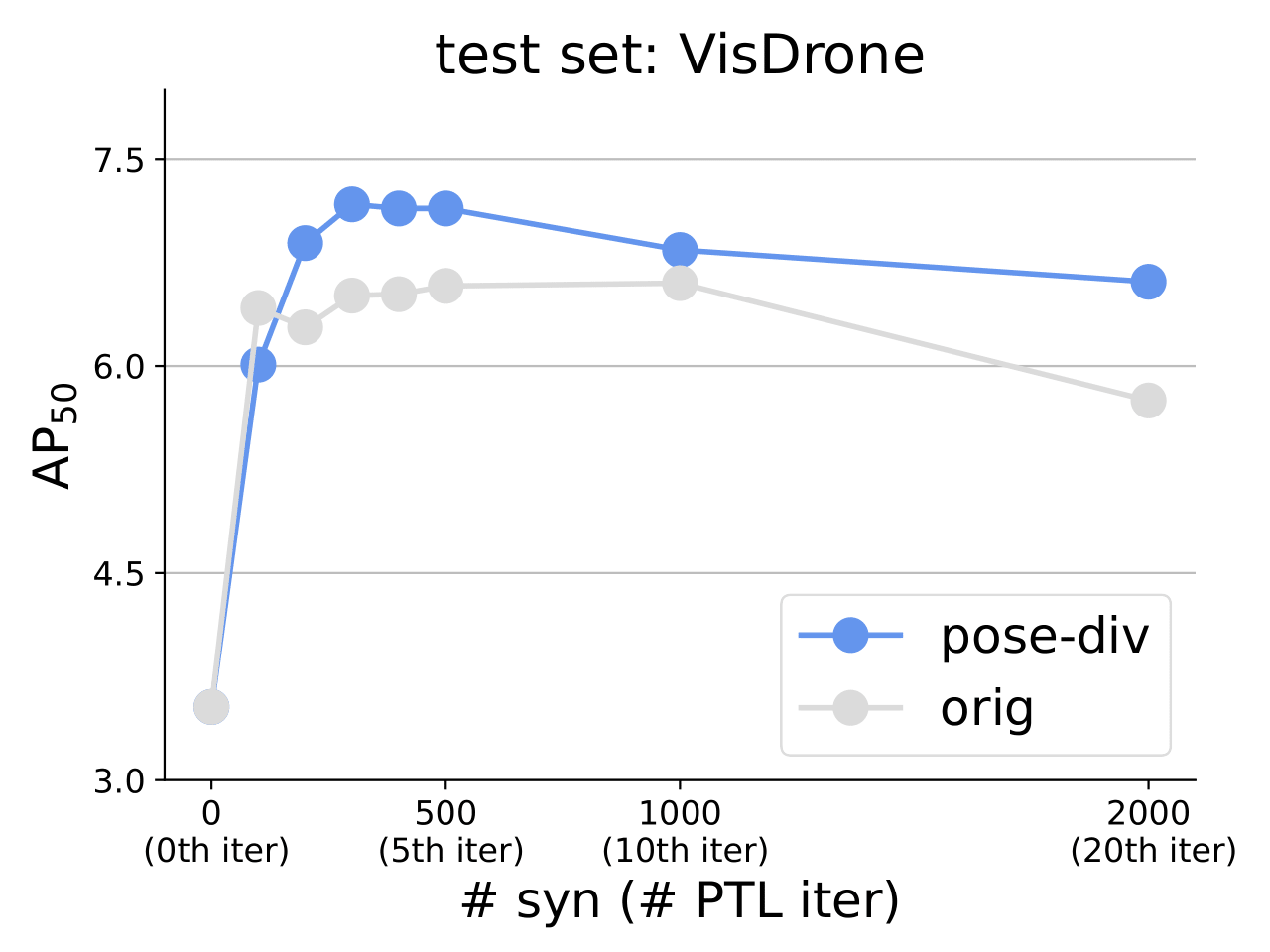} &
\includegraphics[width=.32\linewidth]{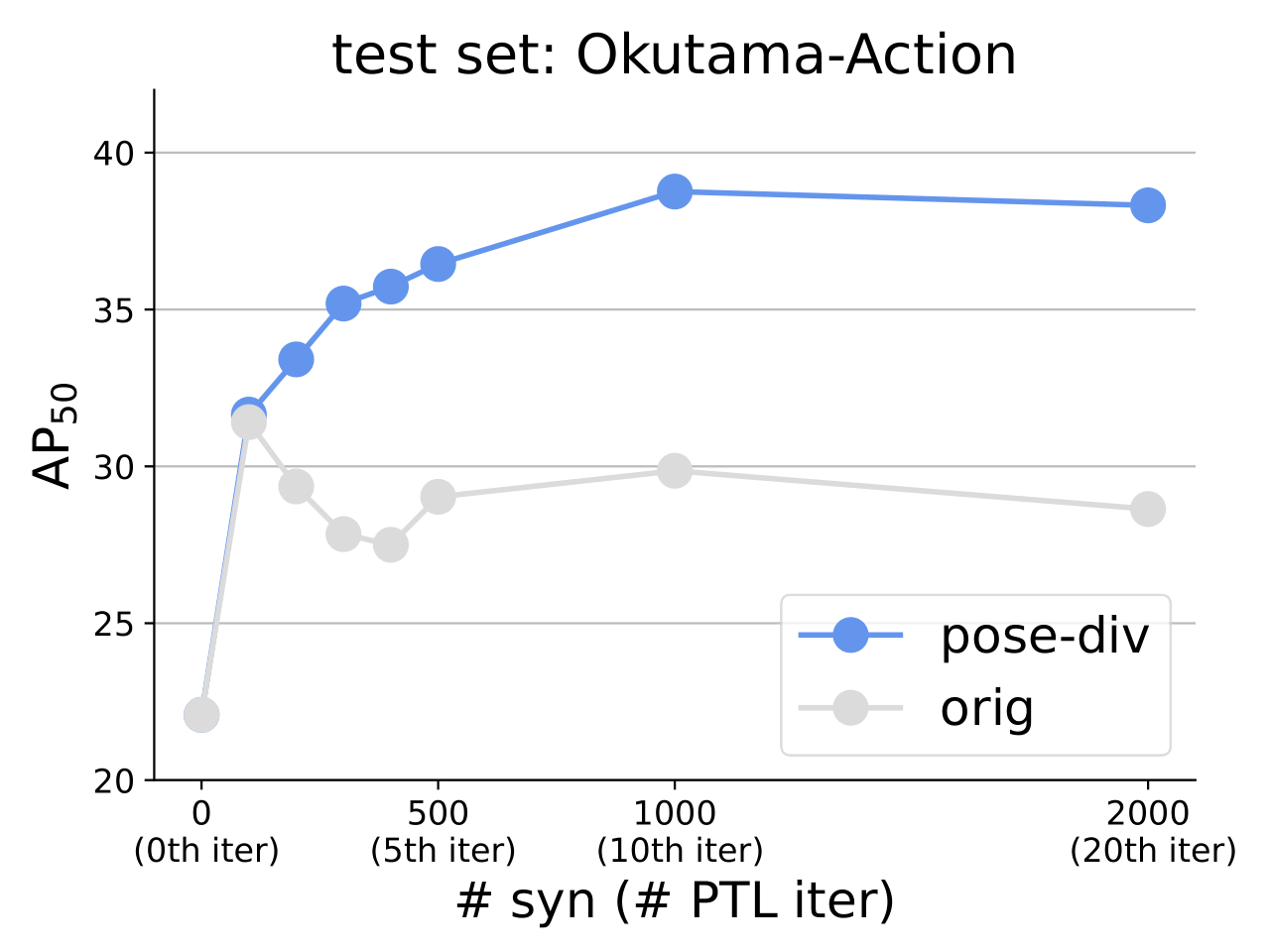} &
\includegraphics[width=.32\linewidth]{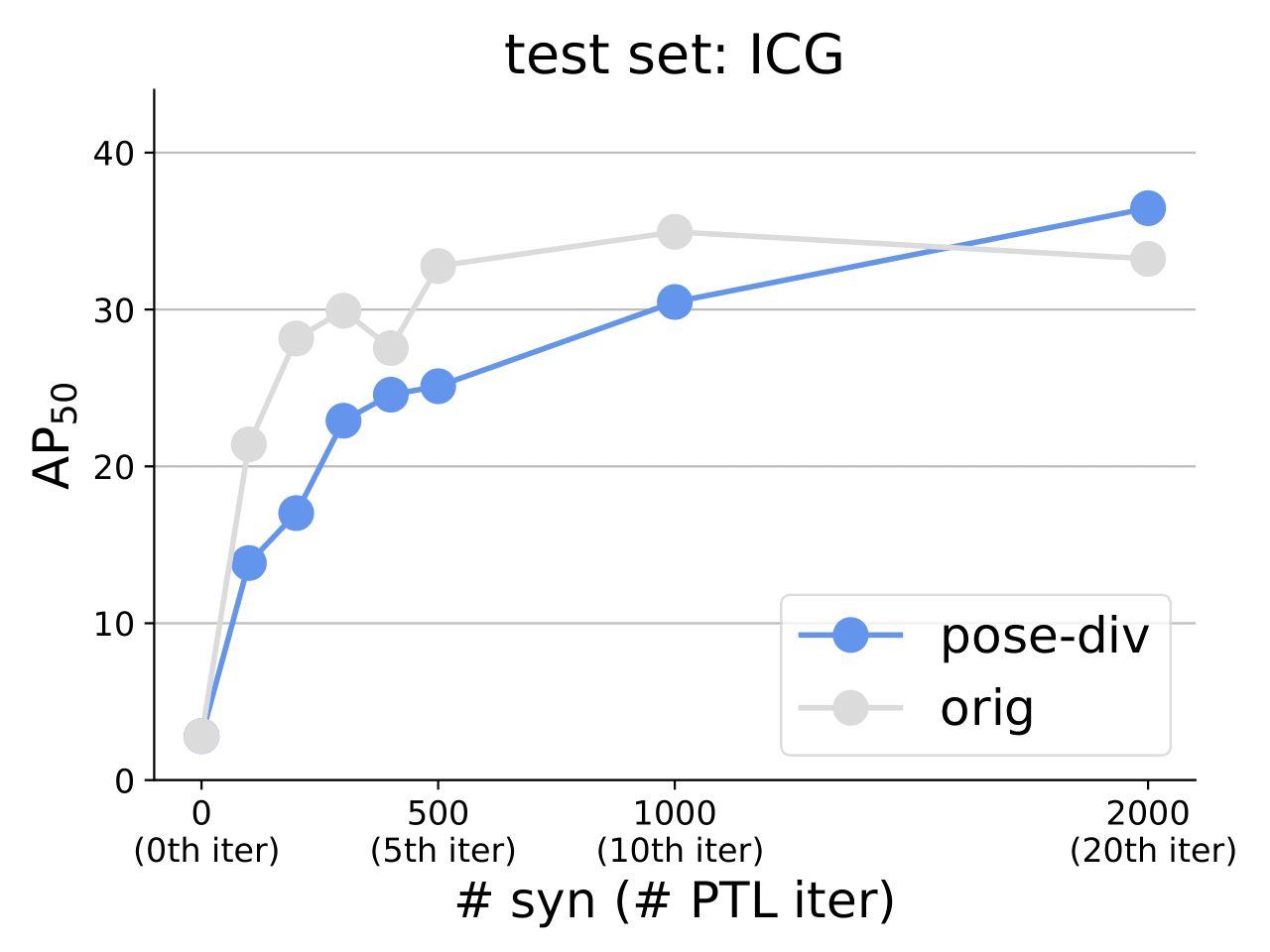} \\
\end{tabular}
\vspace{-0.2cm}
\caption{{\bf Accuracy changes as PTL progresses} under the Vis-20 setup. The scaling behavior is evaluated up to the 20$^\text{th}$ PTL iteration (four times longer than the standard PTL).}
\label{fig:scalability_PTL_iter}
\end{figure}

\noindent{\bf Scaling behavior as PTL evolves.} Fig.~\ref{fig:scalability_PTL_iter} illustrates how PTL gradually increases the synthetic data used during training. Once over 200 synthetic images are selected, a noticeable gap emerges, with the pose-diversified dataset outperforming the original on VisDrone and Okutama-Action. While accuracy for the original dataset plateaus or declines as more images are added, the pose-diversified set continues to improve in cross-domain setups (Okutama-Action and ICG). This highlights \emph{the importance of data diversity in enhancing performance, particularly in cross-domain settings}.\smallskip

\noindent{\bf Scaling behavior of PTL on ICG.} As shown in Fig.~\ref{fig:scalability_PTL_iter}, when 2000 images are selected, the pose-diversified dataset surpasses the original in accuracy on ICG. Initially, performance is lower due to the limited nadir-view images in the pose-diversified set, while ICG mainly contains nadir-view images. However, \emph{increasing data diversity at scale eventually mitigates this limitation and improves accuracy}.\smallskip


\noindent{\bf Conclusion.} We present SynPoseDiv, a framework that modifies synthetic images to adopt novel poses not in the original set. Incorporating  these pose-diversified images into training data enables models to better capture varied human appearances, improving aerial-view human detection. We also examined synthetic dataset scaling, showing that representing diverse human poses enhances performance. For example, with PTL, detection accuracy keeps improving using our pose-diversified set, while it plateaus with the original set.\smallskip

\noindent{\bf Acknowledgements.} This research was sponsored by the Defense Threat Reduction Agency (DTRA) and the DEVCOM Army Research Laboratory (ARL) under Grant No. W911NF2120076. This research was also sponsored in part by the Army Research Office and Army Research Laboratory (ARL) under Grant Number W911NF-21-1-0258. The views and conclusions contained in this document are those of the authors and should not be interpreted as representing the official policies, either expressed or implied, of the Army Research Office, Army Research Laboratory (ARL) or the U.S. Government. The U.S. Government is authorized to reproduce and distribute reprints for Government purposes notwithstanding any copyright notation herein.

\vfill\pagebreak

\bibliographystyle{IEEEbib}
\bibliography{main_ICIP2025}

\end{document}